\documentclass{amia}
\usepackage{lipsum} 

\usepackage{times}
\usepackage{latexsym}
\usepackage{graphicx}
\usepackage{diagbox}
\usepackage{enumitem}
\usepackage{multirow}
\usepackage{indentfirst}
\usepackage{colortbl}
\usepackage{soul}
\usepackage{xcolor}
\usepackage{float}

\newcommand{\hlc}[2][yellow]{{%
    \colorlet{foo}{#1}%
    \sethlcolor{foo}\hl{#2}}%
}

\begin{document}

\title{Context Variance Evaluation of Pretrained Language Models for Prompt-based Biomedical Knowledge Probing}

\author{Zonghai Yao, MSc$^1$, Yi Cao, BS$^1$, Zhichao Yang, MSc$^1$, Hong Yu, PhD$^{1,2,3,4}$}

\institutes{
    $^1$ College of Information and Computer Science, University of Massachusetts Amherst, Amherst, MA, USA; $^2$Department of Computer Science, University of Massachusetts Lowell, Lowell, MA, USA; $^3$Department of Medicine, University of Massachusetts Medical School, Worcester, MA, USA; $^4$Center for Healthcare Organization and Implementation Research, Bedford Veterans Affairs Medical Center, Bedford, MA, USA
}

\maketitle

\section*{Abstract}

\textit{Pretrained language models (PLMs) have motivated research on what kinds of knowledge these models learn. Fill-in-the-blanks problem (e.g., cloze tests) is a natural approach for gauging such knowledge. BioLAMA generates prompts for biomedical factual knowledge triples and uses the Top-k accuracy metric to evaluate different PLMs' knowledge. However, existing research has shown that such prompt-based knowledge probing methods can only probe a lower bound of knowledge. Many factors like prompt-based probing biases make the LAMA benchmark unreliable and unstable. This problem is more prominent in BioLAMA. The severe long-tailed distribution in vocabulary and large-N-M relation make the performance gap between LAMA and BioLAMA remain notable. To address these, we introduced context variance into the prompt generation and proposed a new rank-change-based evaluation metric. Different from the previous known-unknown evaluation criteria, we proposed the concept of "Misunderstand" in LAMA for the first time. Through experiments on 12 PLMs, we showed that our context variance prompts and Understand-Confuse-Misunderstand (UCM) metric make BioLAMA more friendly to large-N-M relations and rare relations. We also conducted a set of control experiments to disentangle "understand" from just "read and copy".
}

\section*{Introduction}

Pre-trained language models (PLMs) like BERT \cite{devlin2018bert} have achieved impressive results on few-shot or zero-shot language understanding tasks by pre-training model parameters in a task-agnostic manner and then fine-tuning to transfer the knowledge to specific downstream tasks \cite{brown2020language, yao2020zero, yao2021improving, kwon2022medjex}. 
Recently, researchers have become interested in measuring how much factual information PLMs get from pre-training. \cite{petroni2019language} formally define this task in the LAMA benchmark, which consists of (subject, relation, object) triples and a corresponding human-written template expressing each relation. They show that BERT can predict objects given task-specific prompts.
In the biomedical domain, \cite{sung2021can} release the Biomedical LAnguage Model Analysis (BioLAMA) probe. They show that biomedical domain specific PLMs like BioBERT \cite{lee2020biobert} can predict objects in biomedical factual knowledge triples given cloze-style prompts—for example, "nasal polyp has symptoms such as [Mask]."

However, existing research has shown that such prompt-based knowledge probing methods can only probe a lower bound of knowledge \cite{petroni2019language, jiang2020can}. 
Even though subsequent work has attempted to improve by finding better prompts \cite{shin2020autoprompt, zhong2021factual}, several biases of the prompt-based probing methods (i.e., Prompt Preference Bias, Instance Verbalization Bias, and Sample Disparity Bias) make the evaluation in the LAMA benchmark unreliable \cite{cao2021knowledgeable, cao2022can}. 
This problem is more severe in N-M relation with large N and M \cite{petroni2019language} and in rare relations exhibiting long-tailed distribution \cite{meng2021rewire}. 
For example, according to Unified Medical Language System (UMLS) \cite{bodenreider2004unified}, asthmatic has 474 symptoms and cough has at least 66 related diseases. And the frequency of cough (the most common symptom of asthmatic) is much more than other rare symptoms.
Since such large N-M relations and rare relations are common in biomedical domain, they make the BioLAMA's performance much lower than LAMA's \cite{cao2021knowledgeable}. 

In this paper, we introduce context variance into the prompt generation to mitigate three prompt-based probing biases highlighted by \cite{cao2022can}. For every biomedical factual knowledge triple, we retrieved contexts from two large and representative biomedical resources: MIMIC-III \cite{johnson2016mimic}, which comprises 59,652 electronic health record (EHR) notes, and a randomly sampled subset of PMC \cite{goldberger2000physiobank}, which includes 50k published biomedical articles. 
Following \cite{yang2020generating} work, we collect a corpus of 50K outpatient EHR notes from the Local Medical Center, named LMC-EHRs dataset. The biomedical PLMs are always pretrained on PubMed and MIMIC-III, we want to create this third context resource for a more fair comparison since none of these PLMs saw these EHR data before.
For a more robust evaluation, we created three different kinds of synthetic contexts, which were generated based on corresponding real contexts from the above three resources, to largely increase the variance setting.
Subsequently, we propose a new rank-change-based evaluation metric, namely Understand-Confuse-Misunderstand (UCM).  Different from the previous known-unknown evaluation criteria in the previous rank-based metric (Top-k ACC), we propose a new concept of "misunderstand" in LAMA for the first time.
Through extensive experiments on 12 PLMs, our context variance prompts and UCM metric makes BioLAMA more friendly in Large-N-M relations and rare relations. In further experiments, we conducted a set of control experiments to disentangle "understand" from just "copy", and showed that PLMs, especially biomedical domain-specific PLMs show evidence of "understanding" of some knowledge. 

    \begin{figure*}[t]
        \centering
        \includegraphics[width=\linewidth]{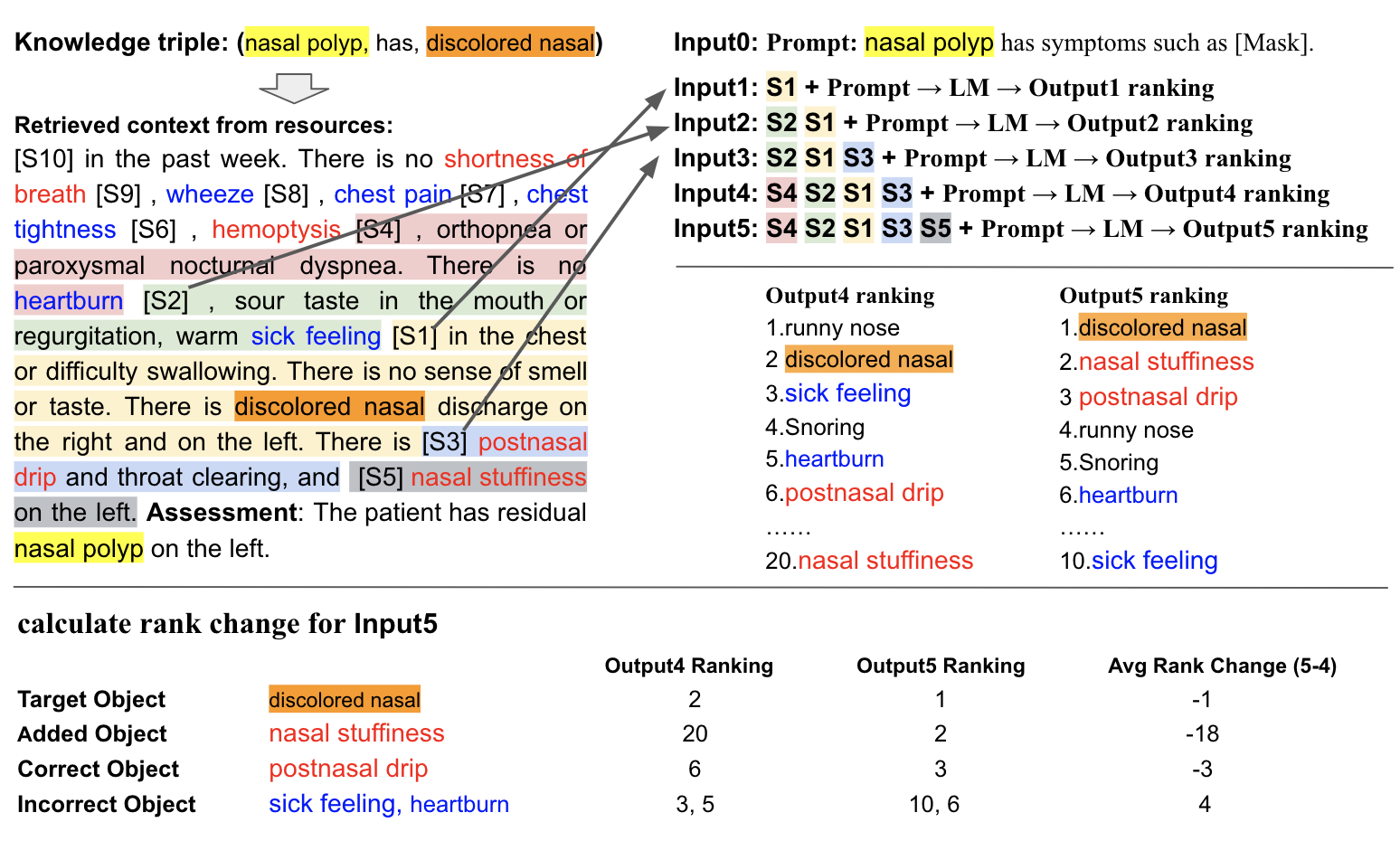}
        \caption{For triple ($S$, relation, $O_t$), \hlc[yellow]{yellow entity} represents $S$, \hlc[orange]{orange entity} represents $O_t$, \textcolor{red}{red entities} represents other entities that are equally correct in the context ($O_{cor}$), and \textcolor{blue}{blue entities} represents wrong entities in the context  ($O_{incor}$). Here, "correct" means that these entities also satisfy the same relation with $S$. We then split the context into different segments according to these entities. Next, we add the context to the prompt segment by segment, and observe that each time a new segment is added, how the ranking of Target Object ($O_t$), Added Object ($O_a$), Correct Object ($O_{cor}$), and Incorrect Object ($O_{incor}$) will change in the "[MASK]" place.}
        \label{UCM_metric}
    \end{figure*}

\section*{Related Work}
\paragraph{Language Models as Knowledge Bases}
\cite{petroni2019language} introduced the LAMA benchmark and began work on prompt-based factual knowledge probing. They pointed out that their benchmark provides only a lower-bound estimate of the amount of factual information stored in PLMs, since their manually written prompts may not be optimal for eliciting facts. Therefore, follow-up work focused on tightening this bound to find more optimal prompts. \cite{jiang2020can, shin2020autoprompt, zhong2021factual, petroni2020context}.
We are also working towards the same goal, but we do it from a different angle. We explore the way of context variance prompting method and re-design its evaluation metrics according to the characteristics of biomedical domain.
Existing work uses Top-k ACC as the evaluation metric, so there are two possibilities for certain knowledge, "Known" (in Top-k) and "Unknown" (not in Top-k). 
\cite{yao2022extracting} added EHR notes as context to the prompt, and proposed a rank-change-base method. They found that the rank-change-base method can also show that PLMs have some knowledge because PLMs have completely different behavior on knowledge and noise under their setting. Our work builds on this rank-change-base method. Thanks to the three situations that will occur in "rank-change" (increase, no change, and decrease), PLMs can have three different status for certain knowledge ("Understand", "Confuse", and "Misunderstand"). Among them, "Misunderstand" is important for the biomedical domain, because the potential harm caused by "Misunderstand", such as hallucination and contradiction, is much greater than just "Unknown" or "Confuse".

\paragraph{BioLAMA} In the biomedical domain, \cite{sung2021can} created and released the BioLAMA benchmark following the LAMA setting, which consists of 49K biomedical factual triples. However, the performance gap between LAMA and BioLAMA remains notable because biomedical domain knowledge probing has its unique challenges (including severe long-tailed distribution in vocabulary, multi-token entities, large N-M relations). \cite{meng2021rewire} proposed a new method, Contrastive-Probe, to handle the multi-token challenge during encoding answers. In this paper, we try to handle the other two challenges, long-tailed distribution in vocabulary and large N-M relations evaluation.

\paragraph{Biases in Prompt-based Knowledge Probing}
\cite{cao2022can} highlighted three critical biases which could impact performance: Prompt Preference Bias (PPB), Instance Verbalization Bias (IVB), and Sample Disparity Bias (SDB). 
PPB shows the LAMA performance may be biased by the fitness of a prompt to PLMs' linguistic preference. which means semantically equivalent prompts may lead to different biased evaluation results.
IVB shows the evaluation results are sensitive and inconsistent to the different verbalizations of the same instance (e.g., representing the COVID-19 with the SARS-CoV-2 or Coronavirus disease 2019). 
SDB shows the performance difference between different PLMs may due to the sample disparity of their pretraining corpora, rather than their ability divergence. 
In this paper, we introduce context variance into the prompt generation to mitigate these three aforementioned biases.


\section*{Methods}
To quantify a model's knowledge, we re-design the BioLAMA evaluation process. Our assumptions are as follows.

\begin{enumerate}[topsep=0.2pt,itemsep=0.2ex,partopsep=.2ex,parsep=0.2ex]
    \item Adding a context containing subject and object and other relevant entities to the prompt can help the BioLAMA task overcome aforementioned challenges and biases.
    \item The previous rank-based metric is not optimal in model stability and reliability. An improved rank-change based metrics, which best captures rank variations between different inputs, can better evaluate PLMs knowledge.
\end{enumerate}

    
\subsection*{Context Segmentation and Retrieval}
\label{Context Retrieval and segmentation}

Assume we have one relation ($Subject$, relation, $Object_{target}$), or ($S$, relation, $O_{t}$), to be evaluated:

\begin{enumerate}[topsep=0.5pt,itemsep=0.2ex,partopsep=0.2ex,parsep=.20ex]
    \item We retrieve documents including both subject $S$ and target object $O_t$ from certain resources. 
    
    \item We will target all other entities that have the same entity-type of $O_t$ (e.g. nasal discharge's entity type is "symptom"), named entity-pool ($O_{pool}$). According to certain Knowledge Bases (KBs) in BioLAMA \cite{bodenreider2004unified, davis2021comparative, turki2019wikidata}, there may be other objects also satisfying the same relation with this subject. So we named all those correct entities in $O_{pool}$ as $O_{cor}$ (red entities in the Figure \ref{UCM_metric}: nasal polyp, symptom, shortness of breath.), and the rest of entities in the $O_{pool}$ as $O_{incor}$ (bleu entities in the Figure \ref{UCM_metric}: nasal polyp, not symptom, wheeze) 
    
    \item We take the $O_t$ as the center and every time look for the next entity in left and right order. We use these entities as delimiters to split the document into segments and ensure that only one entity per segment belongs to $O_{pool}$. These segments are named S1, S2, S3, S4, etc. For example, in Figure \ref{UCM_metric}, the text surrounding the "nasal discharge" is used as S1 until the "sick feeling" is encountered on the left side of S1, and the "postnasal drip" is encountered on the right side of S1, so we ensure that S1 only contains one related entity "nasal discharge". Similarly, S2 adds to the left text of "sick feeling" in the corresponding segment until it encounters "heartburn". S3 adds to the right text of the "postnasal drip" in the corresponding segment until it encounters the "nasal stuffiness". Note that if we can't find a new entity in one side, then we will always search the other side.
 
\end{enumerate}

\subsection*{Context Variance Prompt Generation}
\label{Context Variance Prompt Generation}
We have no restrictions on the prompt generation methods, both the manual prompt and the trained prompt can be combined with our methods for the probing task. Following the BioLAMA work, for certain triple ($S$, relation, $O_t$), we manually generate the corresponding prompts for probing. 

\subsection*{Real Context Prompt}
\label{Real Context Prompt}

In the previous section, we get the required context and split them into several segments. Here, we add these segments to the prompt input in turn and get some new inputs. In order to keep the structure of the context, we will put every segment in the correct position (the same with the original text) when adding a new segment. For example, in Figure \ref{UCM_metric}, we can use the manual prompt method to generate a prompt like "nasal polyp has symptoms such as [Mask].", which is the input0. We put S1 before the prompt and get input1. Because S2 is on the left side of S1 in the original text, so we put S2 before S1 and get input2. Because S3 is on the right side of S1 in the original text, we put S3 after S1 (before the prompt) and get input3, and so on. 

    \begin{figure*}[t]
        \centering
        \includegraphics[width=\linewidth]{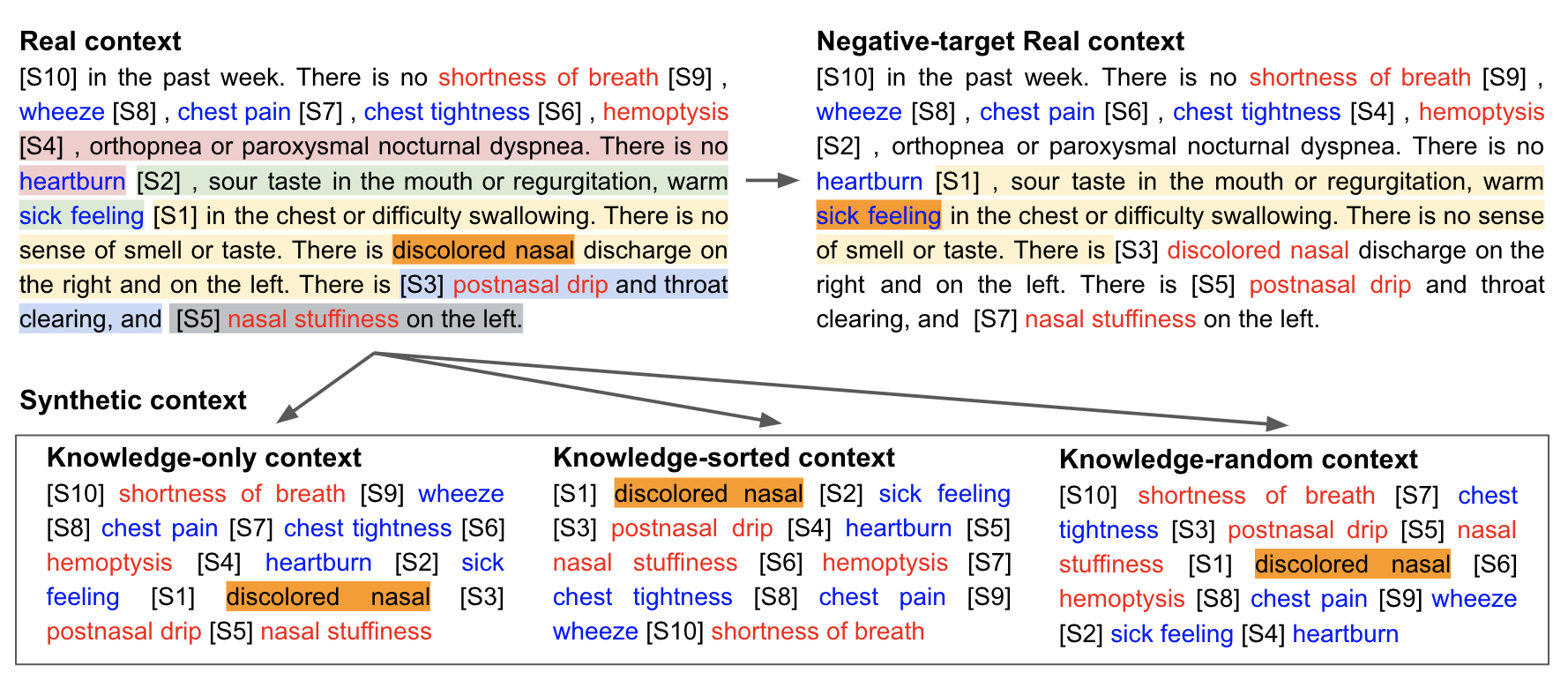}
        \caption{The three Synthetic Contexts and Negative-target Real context are mentioned in the Method section. Assume we already have one Real context, we remove all non-knowledge content of the Real Context to generate the Knowledge-only context. We continue to concatenate S1, S2, S3 ... in the order from left to right to generate the knowledge-sorted context. We randomly shuffle all segments in the Knowledge-only context and then rename S1 to S10 with $O_t$ as the center to generate the Knowledge-random context. For Negative-target Real context, we find the closest $O_{incor}$ to $O_t$ in the Real context and do the segmentation again using this $O_{incor}$ as the new center (the original $O_t$ is only an $O_{cor}$ in the context).}
        \label{context_variance_and_negativeE}
    \end{figure*}

\subsection*{Synthetic Context Prompt}
\label{Synthetic Context Prompt}
We call the contexts described in the previous section as real contexts, because they are retrieved texts from real-world resources. 
However, there are not only knowledge (entities) in the real context, but also other texts. The performance of PLMs in prompt-based probing tasks is not only determined by knowledge, but also affected by language ability in the whole context. In order to measure knowledge more accurately, and also to add more variance, we design three synthetic context generation methods. In Figure \ref{context_variance_and_negativeE}, we continue with the previous example to describe how we generate these synthetic contexts.

\paragraph{Knowledge-only context} For the real context, we remove those non-knowledge texts in each segment, and only keep knowledge in the context. The positions of all segments remain unchanged. The synthetic context generated in this way simply removes the "non-knowledge" content.

\paragraph{Knowledge-sorted context} We sort the segments in ascending order for the Knowledge-only context, and add them to the synthetic context from left to right (S1 S2 S3 ... + prompt). The synthetic context generated here not only removes the "non-knowledge" content, but also changes the relative position between different segments in the input. However, the order in which each segment is added to the input remains unchanged.

\paragraph{Knowledge-random context} We keep S1 in the first because it contains the target object. Then we randomly pick the next segment from the remaining segments and add it to any position in the input. The synthetic context generated in this way only preserves the same object-pool. But the order and position of the segments are added at random.

\subsection*{Negative-target Context}
\label{Negative-target Context}
Because of the attention mechanism, PLMs often "rely too much on copying from the context" \cite{li2019don, petroni2020context}. Specific to behavior, if PLMs do not understand certain knowledge, they may directly copy non-relevant or incorrect entities mentioned in the context. 
Conversely, if PLMs have certain knowledge, they may be less likely to do so.
In other words, in BioLAMA, the knowledgeable PLMs will rank higher
to the correct than incorrect objects in the [MASK] position. 
In order to verify this idea, we designed the corresponding Negative-target context for each context mentioned. As shown in Figure \ref{context_variance_and_negativeE}, we search for $O_{incor}$ from the segments in the context in the order of S1, S2, S3, S4.... We take the first $O_{incor}$ found as the Negative-target of this context, and treat the original $O_t$ as a normal $O_{cor}$. Then we re-construct the context-variance prompt belonging to the negative-target by using all our methods described above with the negative-target as the center. Therefore, for each $O_t$-centered normal data, we will have a corresponding negative-target-centered comparison data. We hope to use such control experiments to explore whether PLMs behave completely different for the two data. We use this as a distinction between "read from context and copy the knowledge" and "really understand the knowledge". 

\subsection*{Rank-change Calculation}
\label{Rank-change Calculation}
As described in the previous section, because of the attention mechanism, PLMs give entities that appear in the context more likely to appear at the [MASK]. Therefore, if we continue to use Top-k ACC, $O_t$ will always have a particularly high probability of being judged as "known", because all the contexts we add contain $O_t$. Since the rank-based metric is not suitable for evaluating the context variance prompt probing task, we design a rank-change-based metric.

As shown in Figure \ref{UCM_metric}, when we have continuous segments S1, S2, S3, S4...., we regard the context as a process of continuously adding new segments centered on $O_t$. In this process, each newly added segment will only contain one entity, named $O_a$. Here, $O_{a}$ may belong to $O_{cor}$, and may belong to $O_{incor}$. Intuitively, PLMs with the corresponding knowledge should have different behaviors for these two different situations. In particular, we try to understand how PLMs behave differently for different entities in context by observing the following four sets of Rank Change (RC) results. As shown in Figure \ref{UCM_metric}, assuming that we add S5 to input4:

\begin{enumerate}[topsep=0.2pt,itemsep=0.2ex,partopsep=.2ex,parsep=0.2ex]
    \item RC-$O_t$: change in the ranking of $O_t$
    \item RC-$O_a$: change in the ranking of $O_a$ in S5
    \item RC-$O_{cor}$: average change in the ranking of all existing $O_{cor}$ in the output4
    \item RC-$O_{incor}$: average change in the ranking of all existing $O_{incor}$ in the output4 
\end{enumerate}

Here, the "change in the ranking" refers to the ranking of the corresponding entity in output5 minus its ranking in output4. The result of RC (output5-output4) will have three cases, ranking-increasing (negative), ranking-unchanged (zero) and ranking-decreasing (positive). It is interesting to observe how PLM's RC behaves differently when a new $O_a$ come ($O_a$ can be either $O_{cor}$ or $O_{incor}$). We put this results in table \ref{behavior}.

\subsection*{Understand-Confuse-Misunderstand Metric (UCM)}
\label{Understand-Confuse-Misunderstand Metric (UCM)}

The BioLAMA tries to verify whether or not a certain PLM has knowledge about ($S$, relation, $O_t$), so in the four groups of RC scores, we can directly take RC-$O_t$ as the metric. 
So, according to the results of Rank-change, we define the UCM metric as follows. Suppose we get a series of RC scores by the method described:

\begin{enumerate}[topsep=0.2pt,itemsep=0.2ex,partopsep=.2ex,parsep=0.2ex]
    \item For each input, we check whether $O_a$ belongs to $O_{cor}$ and leaving only those RC for which $O_a$ belongs to $O_{cor}$
    
    
    \item For the remaining data, we count their distribution, and take the proportion of RC-$O_t$ less than 0 as the Understand score, the proportion of RC-$O_t$ equal to 0 as the Confuse score, and the proportion of RC-$O_t$ greater than 0 as the Misunderstand score
\end{enumerate}

We name the above steps as knowledge-level-UCM ($UCM_k$), which is consistent with the purpose of previous metric Top-k ACC. $UCM_k$ tries to verify whether a certain PLM knows a certain triple (knowledge) in one relation (like disease-symptom relation), and then calculate the respective percentages of U, C, and M for all triples in this relation for evaluation. The previous metric Top-k ACC is doing the same thing, except they compute the percentages of known (in Top-k) and unknown (not in Top-k).

Similarly, if we not only want to verify whether a certain PLM has a certain knowledge, but directly compare whether some PLMs are more knowledgeable, then we can collect all context variance prompts for all relations' triples in a certain KB, and use the same steps above to get the RC distribution over this KB. With this we can compare the UCM scores of different PLMs to tell which PLMs are more knowledgeable. We name it as model-level-UCM ($UCM_m$).

\section*{Results}
\subsection*{Experimental Setup}

We conduct the experiments on 12 PLMs, including BERT (base and large) \cite{devlin2018bert}, RoBERTa (base and large) \cite{liu2019roberta}, BioBERT \cite{ lee2020biobert}, ClinicalBERT \cite{alsentzer2019publicly}, 3 kinds of BioLMs \cite{lewis-etal-2020-pretrained}, and 3 kinds of BlueBERTs \cite{peng2019transfer}. All the results in the Experiments part are the average scores of three different resource data as context.
We use manual prompts created by domain experts following BioLAMA. It is important to note that we have no restrictions on the prompt generation methods, both the manual prompt and the trained prompt can be combined with our methods. 
We use a fill-in-the-blank cloze statement (i.e., a "prompt") for probing. Since the majority of entities in our dataset are made up of multiple tokens, we also implement a multi-token decoding strategy following \cite{sung2021can}.
We mainly use UCM introduced our Method section as the evaluation metric, but we also report the results of using the previous metrics following BioLAMA with Top-k accuracy, which is 1 if any of the Top k object entities are included in the annotated object list, and is 0 otherwise

\subsection*{Rank-change} 

    \begin{table}
        \centering
        \begin{tabular}{l|cccc|cccc}
        \hline
        
        & \multicolumn{4}{c|}{$O_a \in O_{cor}$} & \multicolumn{4}{c}{$O_a \in O_{incor}$} \\
        
        & $O_t$ & $O_a$ & $O_{cor}$ & $O_{incor}$ & $O_t$ & $O_a$ & $O_{cor}$ & $O_{incor}$\\
        
        \hline
        BERT & -0.98 & -6.56 & -2.8 & 0.045 & 0.78 & -1.98 & 0.74 & -0.59\\
        
        BioBERT & -2.07 & -12.7 & -5.62 & 0.23 & 1.04 & -6.05 & 1.06 & -2.02\\
        
        BioLM & -1.59 & -10.56 & -5.2 & 0.09 & 0.88 & -3.76 & 0.91 & -1.32\\

        \hline
        \end{tabular}
        \caption{With different contexts, rank-change measures changes of predictions of [MASK] in prompts by different PLMs. There are three cases for rank change, ranking-increasing (negative), ranking-unchanged (zero) and ranking-decreasing (positive).
        }
        \label{behavior}
    \end{table}

Table \ref{behavior} shows PLMs' rank-change behaviors. Generally speaking, because $O_a$ is added to the context, there will be a ranking-increase because of the attention mechanism. $O_t$ belongs to $O_{cor}$, so intuitively there will be the same trend of RC. And $O_{cor}$ and $O_{incor}$ should have the completely opposite trends. Depending on the difference of $O_a$ in the newly added segment ($O_a$ may belong to either $O_{cor}$ or $O_{incor}$), the specific changes of the four groups of RC will also be different. We find that if $O_a$ belongs to $O_{cor}$, then $O_{cor}$ and $O_t$ should have the same trend of change (rank increase) as $O_a$, otherwise $O_{incor}$ will have the same trend of change as $O_a$.
    
\subsection*{$UCM_K$ and Top-k ACC}
\label{Compare UCM_K and Top-k ACC}

Here we show that UCM can be a more appropriate metric than Top-k ACC. We compare $UCM_k$ with top-k accuracy with or without context.Table \ref{UCM_k_ACC_wo_context_comparasion} shows that evaluation results with Top-k ACC is often counter-intuitive and unstable. For example, PLMs pretrained on the biomedical domain corpus (e.g., clinicalBERT) get much lower scores than the general domain PLMs (e.g., BERT). The clinicalBERT and BioBERT are trained in the biomedical domain and have similar performance in most of the downstream tasks, yet they show a very huge gap in their performance.
Table \ref{UCM_k_ACC_wo_context_comparasion} shows that when context is included, almost all BioLMs' ACC@5 have non-distinguishable ~90\% scores,
and we cannot distinguish whether the model is 'copy' or understand based on the evaluation with Top-k ACC.

In contrast, the results with $UCM_k$ are more interpretable and consistent. For both BERT families and BioLMs families in table \ref{UCM_k_ACC_wo_context_comparasion},
the results show that large models have better scores (higher U, lower or similar C and M) than the corresponding base models. We additionally analyze the scores for each resource. We find that when we use LMC-EHRs data as the context, biomedical PLMs (BioBERT, clinicalBERT, RoBERTa-base-PM, RoBERTa-base-PM-M3) all have better performance (higher U, lower C and M) than the general domain PLMs (BERT-base and RoBERTa-base). Since MIMIC-III is an EHR resource and in constrast PMC comprises biomedical literature articles, PLMs pretrained on MIMIC-III have a better performance (similar U and C, lower M) on LMC-EHRs data. All these results illustrates that introduction of "Misunderstand" makes the results more consistent and interpretable. Although PLMs trained on both PMC and MIMIC-III have knowledge inside LMC-EHRs context, a more similar domain (MIMIC-III) can provide more accurate knowledge, making these PLMs less "Misunderstand".

    \begin{table}
        \centering
        \begin{tabular}{l|cc|cc|ccc}
        \hline
        
        & \multicolumn{2}{c|}{without context} & \multicolumn{5}{c}{with context} \\
        
        \hline
        
        &  \small{acc@1} &  \small{acc@5} & \small{acc@1} &  \small{acc@5} & U & C & M\\
        
        \hline
        \small{BERT-base} & \small{0.0038} & \small{0.57} & \small{0.327} & \small{0.68} & \small{0.119} & \small{0.787} & \small{0.094} \\
        
        \small{BioBERT} & \small{0.227} & \small{0.639} & \small{0.407} & \small{0.881} & \small{0.328} & \small{0.427} & \small{0.243} \\
        
        \small{ClinicalBERT} & \small{0.014} & \small{0.264} & \small{0.391} & \small{0.812} &  \small{0.302} & \small{0.533} & \small{0.165} \\
        
        \small{BERT-large} & \small{0} & \small{0.009} & \small{0.258} & \small{0.665} & \small{0.158} & \small{0.738} & \small{0.103}\\
        
        \hline
        
        \small{RoBERTa-base} & \small{0.11} & \small{0.218} & \small{0.572} & \small{0.872} & \small{0.287} & \small{0.495} & \small{0.217}\\
        
        \small{RoBERTa-base-PM} & \small{0.134} & \small{0.281} & \small{0.412} & \small{0.902} & \small{0.298} & \small{0.427} & \small{0.274}\\
        
        \small{RoBERTa-base-PM-M3} & \small{0.102} & \small{0.273} & \small{0.563} & \small{0.904} & \small{0.299} & \small{0.432} & \small{0.267}\\
        
        \hline
        
        \small{RoBERTa-large} & \small{0.107} & \small{0.229} & \small{0.503} & \small{0.865} & \small{0.304} & \small{0.509} & \small{0.186}\\
        
        \small{RoBERTa-large-PM-M3} & \small{0.105} & \small{0.31} & \small{0.644} & \small{0.915} & \small{0.312} & \small{0.432} & \small{0.255}\\
        
        \hline
        \end{tabular}
        \caption{Top-k ACC VS $UCM_k$}
        \label{UCM_k_ACC_wo_context_comparasion}
    \end{table}

    \begin{table}
        \centering
        \begin{tabular}{l|ccc|ccc}
        \hline
        
        & \multicolumn{3}{c|}{target} & \multicolumn{3}{c}{negative-target} \\
        
        & U & C & M & U & C & M\\
        
        \hline
        \small{BERT-base} & \small{0.205} & \small{0.73} & \small{0.060} & \small{0.026} & \small{0.92} & \small{0.049}\\
        
        \small{BioBERT} & \small{0.32} & \small{0.584} & \small{0.095} & \small{0.103} & \small{0.72} & \small{0.175}\\
        
        \small{ClinicalBERT} & \small{0.37} & \small{0.56} & \small{0.061} & \small{0.07} & \small{0.79} & \small{0.134}\\
        
        \small{RoBERTa-base} & \small{0.23} & \small{0.67} & \small{0.099} & \small{0.055} & \small{0.83} & \small{0.11}\\
        
        \small{RoBERTa} & \small{0.301} & \small{0.583} & \small{0.115} & \small{0.109} & \small{0.72} & \small{0.168}\\
        
        \small{RoBERTa-base-PM-M3} & \small{0.335} & \small{0.55} & \small{0.112} & \small{0.107} & \small{0.77} & \small{0.12}\\
    
        \small{BlueBERT-base-PM} & \small{0.383} & \small{0.554} & \small{0.062} & \small{0.026} & \small{0.89} & \small{0.085}\\
        
        \small{BlueBERT-base-PM-M3} & \small{0.407} & \small{0.559} & \small{0.033} & \small{0.025} & \small{0.89} & \small{0.078}\\
        
        \hline
        
        \small{BERT-large} & \small{0.199} & \small{0.75} & \small{0.045} & \small{0.026} & \small{0.9} & \small{0.047}\\
        
        \small{RoBERTa-large} & \small{0.325} & \small{0.614} & \small{0.061} & \small{0.047} & \small{0.86} & \small{0.113}\\
        
        \small{RoBERTa-large-PM-M3} & \small{0.346} & \small{0.556} & \small{0.097} & \small{0.063} & \small{0.82} & \small{0.113}\\
        
        \small{BlueBERT-large-PM-M3} & \small{0.388} & \small{0.504} & \small{0.108} & \small{0.053} & \small{0.85} & \small{0.095}\\

        \hline
        \end{tabular}
        \caption{PLMs' $UCM_m$ scores of $O_t$ as center or negative-target as center}
        \label{UCM_m}
    \end{table}

\subsection*{Evaluation with UCM}
\label{Evaluation_UCM}

We compute $UCM_k$ and $UCM_m$ to compare the knowledge level between different PLMs. We want to note that even though we show that UCM is a more reliable metric than Top-k ACC, it is not easy to fairly compare the knowledge level between different PLMs. Too many factors may affect the knowledge learned by the PLMs during pretraining, so we have to choose the more easily interpretable models to compare. For example, in table \ref{UCM_k_ACC_wo_context_comparasion} and \ref{UCM_m}, RoBERTa-base is a general-domain PLM, RoBERTa-base-PM continues to train RoBERTa-base on PubMed, and RoBERTa-base-PM-M3 continues to train RoBERTa-base-PM on MIMIC-III. So the comparison between this group of PLMs is more meaningful, because the results in table \ref{UCM_k_ACC_wo_context_comparasion} and \ref{UCM_m} can tell us that if the PLMs continue to train on the target-domain related data, their knowledge level will be higher. 

According to table \ref{UCM_m}, We have the following findings based on the $UCM_m$ scores. We re-emphasize that although the findings we concluded from the $UCM_m$ results are not particularly surprising, such "reasonable" and "intuitive" results can not be obtained with previous metrics. So we make up for these evidences for the biomedical domain with a more reasonable BioLAMA metric.

\begin{enumerate}[topsep=0pt,itemsep=0ex,partopsep=0ex,parsep=0ex]
    \item Large models are always better
    
    \item The similarity between the context domain and PLMs' pretraining domain is decisive 
    
    \item Training longer does not make PLMs more knowledgeable, which is in line with recent work \cite{goyal2021training}
    
    
    \item Real and Knowledge-only contexts do not have much differences. 
    For three synthetic contexts, Knowledge-only context and Knowledge-sorted context do not have big differences, but the Knowledge-random context results are unstable. 
    
\end{enumerate}

\section*{Discussion}

\subsection*{Copy or Understand}

The attention mechanism makes PLMs tend to copy knowledge entities in context. The Top-k ACC results with and without context in table \ref{UCM_k_ACC_wo_context_comparasion} also prove that this phenomenon does exist in our task setting. So, after adding knowledge into context prompts, there is a possibility that PLMs copy from context without understanding. To distinguish the two, we introduce the negative-target context in the Method section. In table \ref{UCM_m}, by comparing PLMs' $UCM_m$ score $O_t$ as center or instead negative-target as center, we find that the behaviors of PLMs for the two are almost completely opposite. This control experiment suggests that PLMs understand these knowledge, and do not just copy.

\subsection*{N-M Relations and Rare Relations Friendly}

The performance gap between LAMA and BioLAMA remains notable because biomedical domain knowledge probing has its unique challenges, and as stated previously, two of them are severe long-tailed distribution in vocabulary (rare relation unfriendly) and many large-N-M relations \cite{sung2021can, meng2021rewire}. Because the rare relation rarely appears in pretraining data, it is difficult for prompts to elicit these facts. The reason behind this is that PLMs are context sensitive, and only specific prompts as contexts allow them to generate corresponding results. For common relation, they are frequently mentioned in different contexts in pretraining, so the requirements for prompt are relatively broad. In contrast, even if a PLM knows a rare relation, it has very strict requirements on the prompt, that is, the prompt used must be very related to the context in these PLMs' pretraining step. On the other hand, the large-N-M relations means the prompt has lots of objects that may be the right answers, however, the model cannot distinguish the difference between $O_t$ and other $O_{cor}$ in the Figure \ref{UCM_metric}. Due to the softmax bottleneck of PLMs \cite{yang2017breaking, chang2022softmax}, if the model prefers to put other $O_{cor}$ on Top-k, it may not be able to put $O_t$ on Top-k at the same time. Therefore, the previous evaluation is unfair and inaccurate.
In this paper, we introduce context variance in the prompt generation process to alleviate the above two problems. Because of the attention mechanism, PLMs often "rely too much on copying from the context" \cite{li2019don, petroni2020context}, which means PLMs usually directly copy other entities mentioned in the context. This is sometimes a big weakness of PLMs for generational tasks, but it is unexpectedly well suited for our use case. If we add the context containing a specific set of $O_{pool}$ to the prompt as the input, PLMs will naturally have candidates (i.e. $O_{pool}$) at the [MASK] position because of attention mechanism. As long as candidates contain $O_t$, no matter whether ($S$, relation, $O_t$) is rare or common relations, or 1-1 or N-M relations, we can use the unified method to observe different behaviors when PLMs treat different inputs. We define a set of procedures in the next few sections to determine whether PLMs know this knowledge according to different behaviors.

\subsection*{Mitigating PPB, IVB, and SDB Biases}
\cite{cao2022can} highlighted three critical biases which affect the correctness and fairness of evaluating different PLMs on the BioLAMA benchmark. We evaluate whether context variance and UCM can alleviate the three biases. 
Here we discuss how the context variance prompts mitigate these biases.
We control Prompt Preference Bias (PPB) by adding context variance, although the prompt itself does not change, the input to the PLMs is not the same. Because UCM is a statistical result of a large number of context variance prompts, only those PLMs that can be regarded as "Understand" in most data are treated as "really know" the corresponding knowledge. Therefore, UCM will not encounter serious PPB like the previous method because only one prompt is used for probing all PLMs.
For Instance Verbalization Bias (IVB), we first map text to the UMLS concepts and therefore eliminate the verbalization challenge at the method level. At the same time, the context variance prompts contain a large amount of variations from real-world resources, so different verbalization can be used as input to PLMs. These all help us relieve IVB.
We control Sample Disparity Bias (SDB) by choosing resources more fairly. All biomedical domain specific PLMs selected in this paper are pre-trained on the representative biomedical text resources: PubMed or MIMIC-III. We also use LMC-EHRs as a third resource, and no PLMs in our evaluation have seen this dataset.

\subsection*{Conclusion}

In this paper, we show that adding context containing relevant entities to the prompt can mitigate two unique challenges (long-tailed distribution in vocabulary and many large-N-M relations) and three biases (PPB, IVB, and SDB) in the BioLAMA task. Then we find the previous rank-based metric is not optimal in model stability and reliability. An improved rank-change-based metric, which best captures rank variations between different inputs, can better evaluate PLMs' knowledge.
We hope that the evidence in our paper will inspire the BioNLP community to pay more attention to what kinds of medical knowledge nowadays language models can learn.

\makeatletter
\renewcommand{\@biblabel}[1]{\hfill #1.}
\makeatother

\bibliographystyle{vancouver}
\bibliography{amia}  

\end{document}